\documentclass{article}

\usepackage{PRIMEarxiv}
\usepackage[utf8]{inputenc} 
\usepackage[T1]{fontenc}    
\usepackage{hyperref}       
\usepackage{url}            
\usepackage{booktabs}       
\usepackage{amsfonts}       
\usepackage{nicefrac}       
\usepackage{microtype}      
\usepackage{lipsum}
\usepackage{fancyhdr}       
\usepackage{graphicx}       
\graphicspath{{media/}} 
\usepackage{amsmath}

\title{Retrieval Augmented Generation and Representative Vector Summarization for Large Unstructured Textual Data in Medical Education
}

\author{
  S. S. Manathunga \\
  Department of Pharmacology \\
  University of Peradeniya \\
  Sri Lanka\\
  \texttt ssm123ssm@gmail.com \\
\And
  Y. A. Illangasekera \\
  Department of Pharmacology \\
  University of Peradeniya \\
  Sri Lanka\\
  \texttt yasi04@gmail.com \\
}

\DeclareMathOperator*{\argmin}{argmin}
\usepackage[normalem]{ulem}
\useunder{\uline}{\ul}{}

\begin{document}

\maketitle

\begin{abstract}
 Large Language Models are increasingly being used for various tasks including content generation and as chatbots. Despite their impressive performances in general tasks, LLMs need to be aligned when applying for domain specific tasks to mitigate the problems of hallucination and producing harmful answers. Retrieval Augmented Generation (RAG) allows to easily attach and manipulate a non-parametric knowledgebases to LLMs. Applications of RAG in the field of medical education are discussed in this paper. A combined extractive and abstractive summarization method for large unstructured textual data using representative vectors is proposed.
 
\end{abstract}


\section{Introduction}

Large language models (LLMs) are demonstrating amazing zero-shot learning capabilities. They have many applications including content generation and general question answering. These models learn the data they were trained on very well and the knowledge is retained encoded in the model weights as a parameterized knowledgebase. This also makes it difficult to revise or update the knowledge of an LLM later, once it is trained on a large corpus of training data \cite{singhal_large_2022}. The models can hallucinate, producing factually inaccurate, plausible-sounding answers. Retrieval Augmented Generation (RAG) tries to mitigate these issues by attaching a non-parametric knowledgebase to the LLM, in the form of a vector database \cite{lewis_retrieval-augmented_2021}. Knowledge in the vector database can easily be modified and different kinds of unstructured textual data can be vectorized and stored in a single vector database.  

RAG-assisted summarization is an important aspect of document intelligence as well. It poses more challenges than retrieval. A well-written summary should highlight the key points and provide an overview of the document's topics. Generating summaries of large documents directly using LLMs is difficult because the input content is often too large to fit into the model's context window. Performance tends to decrease with increasing context lengths even in large context window models. Furthermore, the 'Lost in the Middle' problem, the phenomenon that the LLMs tend to ignore facts in the middle of the context and assign more importance to the beginning and the end of the context becomes worse with large input contexts \cite{johnson_billion-scale_2017}.

To overcome these challenges, RAG-assisted Representative Vector Summarization (RVS) is introduced in this paper. RVS selects a pre-defined number ($k$) of representative text chunks from the non-parametric knowledgebase and applies a combined abstractive and extractive summarizing workflow to generate the final summary. The value of parameter $k$ is determined based on the maximum token limit that can be afforded. The model can identify keywords of the text chunks and their relative importance and create visual representations of the overall content of the document using word clouds and scatter plots.

RVS is implemented in \verb+docGPT+, a document intelligence program written in Python using \verb|langchain| framework and the source is available at \url{https://github.com/ssm123ssm/docGPT-pharm} \cite{chase_langchain_2022}.

\section{Methods}
\label{sec:headings}
Described here is the workflow implemented in docGPT.

\subsection{Retrieval}

\begin{figure}
  \centering
  \fbox{\includegraphics[width=0.9\textwidth]{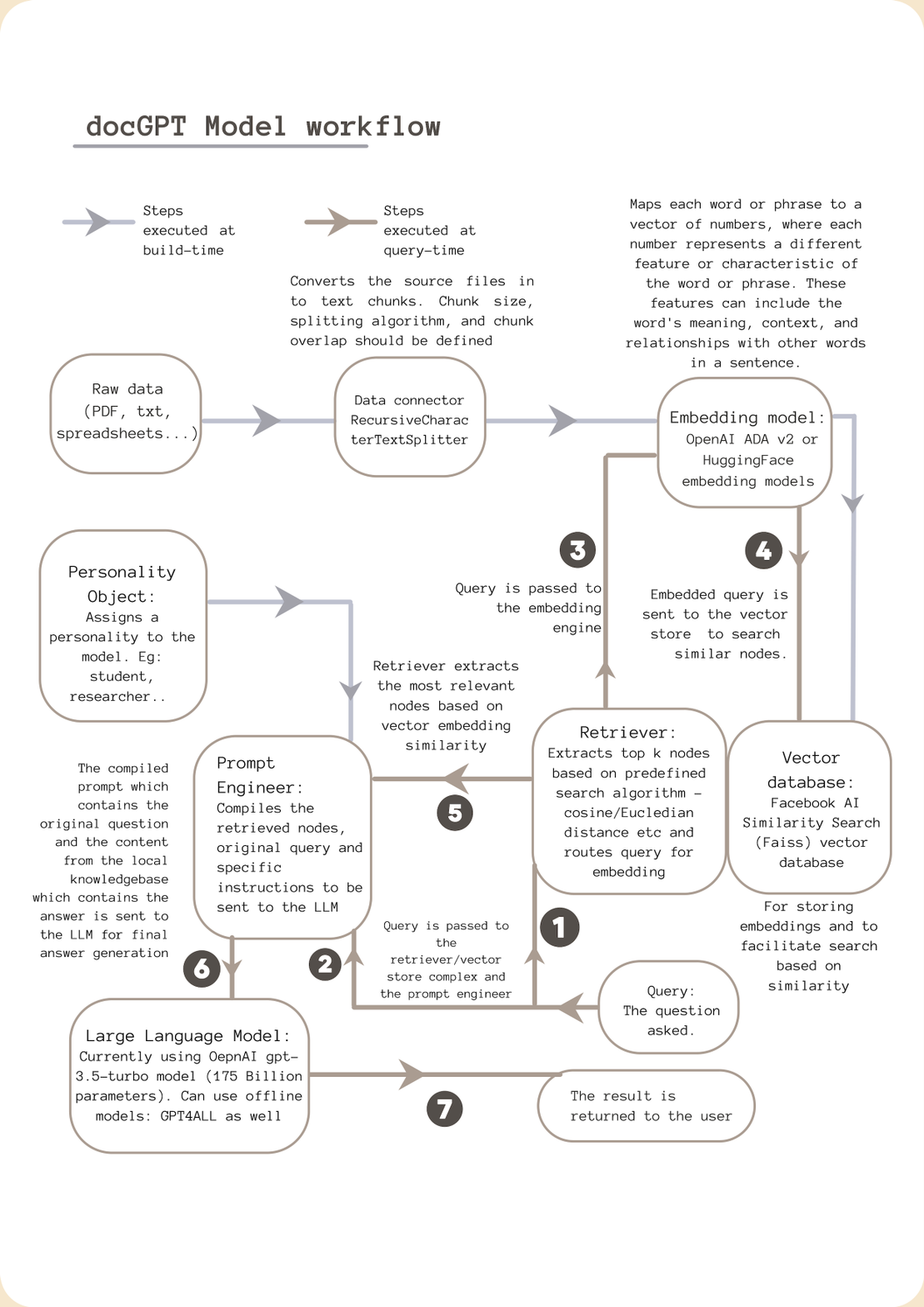}}
  \caption{Retrieval workflow of docGPT}
  \label{fig:fig1}
\end{figure}

The model extracts text from unstructured sources including PDFs, text documents, spreadsheets and slide presentations. It can extract text from images and scanned PDF files using optical character recognition (OCR) \cite{smith_overview_2007}. The extracted text is initially stored in memory as a Document and then the Document is split into text chunks using recursive character text splitting. The chunks are then embedded in a 1536-dimensional vector space using OpenAI \verb|text-embedding-ada-002| embedding engine and stored in Facebook AI Similarity Search (\verb|FAISS|) vector database \cite{johnson_billion-scale_2017}. The \verb|FAISS| vector database is used to find the $k$-most similar chunks to a given query at the query time. The original query, combined with the retrieved chunks is compiled into a prompt and passed to the LLM for generating the answer. The model workflow and the steps excecuted at build-time and at query-time are depicted in Figure \ref{fig:fig1}.

\subsection{Summarization}

\subsubsection{Representative Vector }

A maximum affordable token limit ($T$) is initially defined. The target is to select $k$ number of chunks which is the maximum number of chunks that can be selected without their total token size exceeding $T$, from the vector database of $n$ chunks. Assuming each chunk has an average token size of $s$, $k$ is simply obtained by optimizing for $\max_{k \leq n, k \times s \leq T} k$

Once $k$ is calculated, the chunks in the high-dimension vector space are quantized using the k-means clustering algorithm. $k$-number of clusters which minimize the within-cluster sum of squared Euclidian distances from the corresponding centroids are created within the vector space. Since the distribution of chunks in the vector space is based on their contextual similarity, we assume each cluster captures different aspects/semantics of the original document. 

After this quantization, one representative chunk, which is the closest to the corresponding centroid from each cluster is extracted.  

Let vectors be an  $n\times d$ matrix, where $n$ is the number of chunks and $d$ is the dimensionality of the embeddings. Let $centroids[i]$ denote the centroid of $i$-th cluster. The Euclidian distance between the $m$-th data point and the centroid of the $i$-th cluster is calculated by

\begin{equation}
    distance_m = \sqrt{\sum_{j=1}^d (vectors[m][j] - centroids[i][j])^2}
\end{equation}

and stored in the $distances[i]$ array.

Next, the index of the data point that has the minimum distance to the centroid of the $i$-th cluster, denoted as $closest\_index_i$ , is calculated by:

\begin{equation}
    closest\_index_i = \argmin{distances_i}
\end{equation}

Finally, the $closest\_index_i$ values for all clusters are stored in a list.

$closest\_indices=[closest\_index_0,closest\_index_1,\dots,closest\_index_{k-1}]$

Once the indices of the representative chunks are identified, they are stored in a separate representative Document list.

\subsubsection{Keyword generation and mapping}

Even though the token size of the representative text chunks obtained with this method is relatively smaller compared to the original document, it can still be too large to fit into the LLM's context window. For example, \verb|docGPT| uses a default maximum affordable token limit of 10,000 tokens. The default LLM that \verb|docGPT| calls is OpenAI \verb|gpt-3.5-turbo|, which has a maximum context window of 4k. Therefore, the representative Document list undergoes an intermediate step of extractive summarization (keyword generation and mapping) for each chunk in it.

Three keywords are generated for each representative chunk and the keywords are distributed among all the other members of the same cluster. A word cloud is generated from all the keywords from the n chunks. The relative frequency that each keyword appears is reflected by the size of the word in the word cloud.

The original high-dimensional vectors are then reduced to two dimensions using t-distributed Stochastic Neighbor Embedding (t-SNE) and plotted on a scatter plot with colors corresponding to their clusters and keywords. This enables the identification of the distribution of the contents of the entire document at a glance.

Finally, the mapped summaries are used to create a final abstractive summary and to create a list of key points from the documents.

\section{Evaluation}
\label{sec:others}

\subsection{Retrieval}

Responses for queries on clinical medicine and general pharmacology from LLM without a non-parametric knowledgebase and a RAG model with vector databases built from Kumar and Clark's Clinical Medicine 10th Edition and British National Formulary 82 were compared and checked for accuracy \cite{noauthor_kumar_nodate, noauthor_british_nodate}. The responses generated by \verb|chatGPT| and RAG implementation in \verb|docGPT| are tabulated with the excerpts from the reference books in the supplementary material named Queries. For each query, \verb|docGPT| generated more targetted and accurate answers, while \verb|chtGPT| answers were more generic.  

\subsection{Summarization}

Performances of RVS and retrieve-stuff-summarize methods were tested using Kumar and Clark Clinical Medicine 10th edition and BNF 82 ebooks. Results are summarized in supplementary material named Summaries.

\begin{figure}
  \centering
  \fbox{\includegraphics[width=0.5\textwidth]{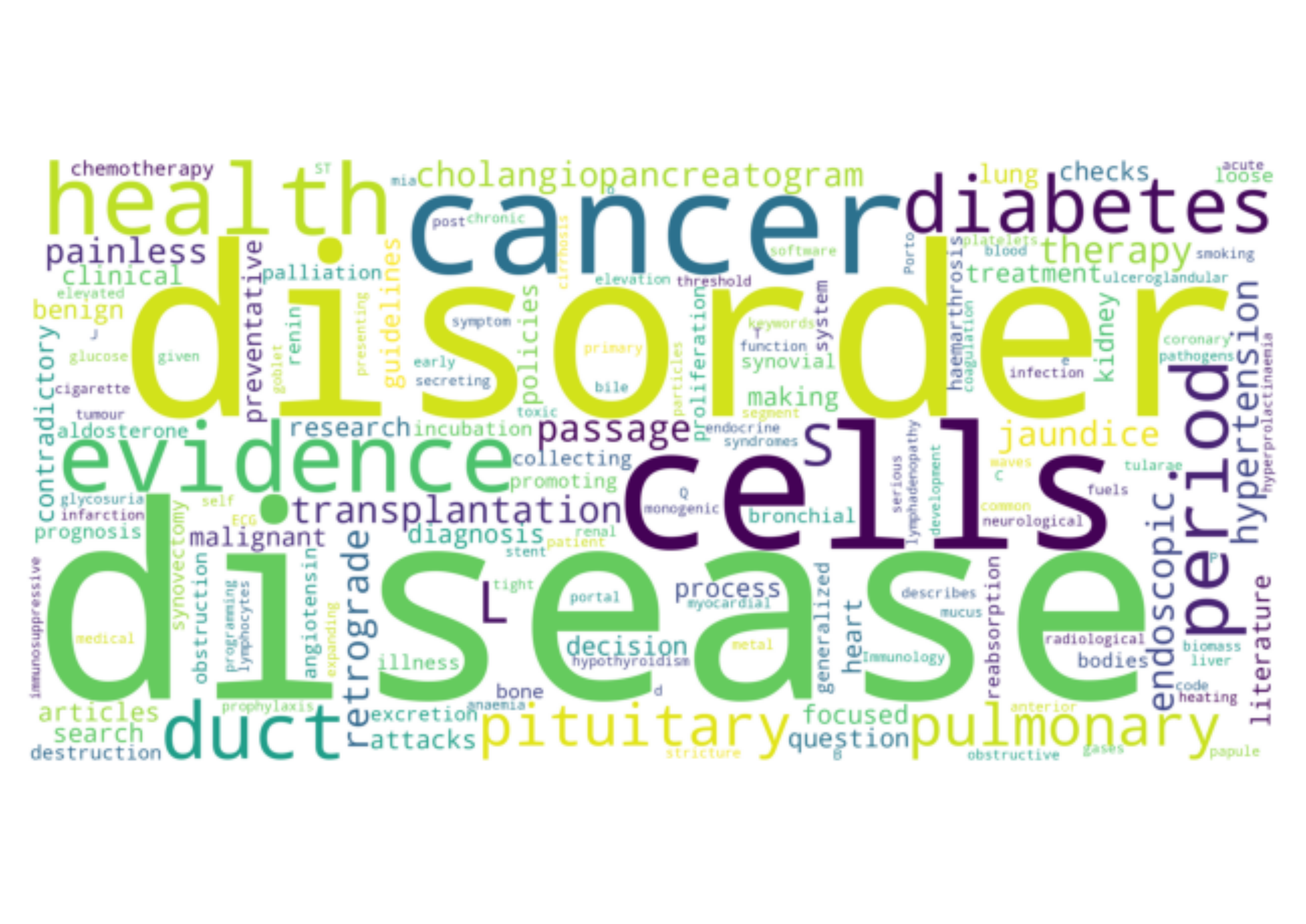}}
  \caption{Word cloud for Kumar and Clark Clinical Medicine}
  \label{fig:fig2}
\end{figure}

\begin{figure}
  \centering
  \fbox{\includegraphics[width=0.5\textwidth]{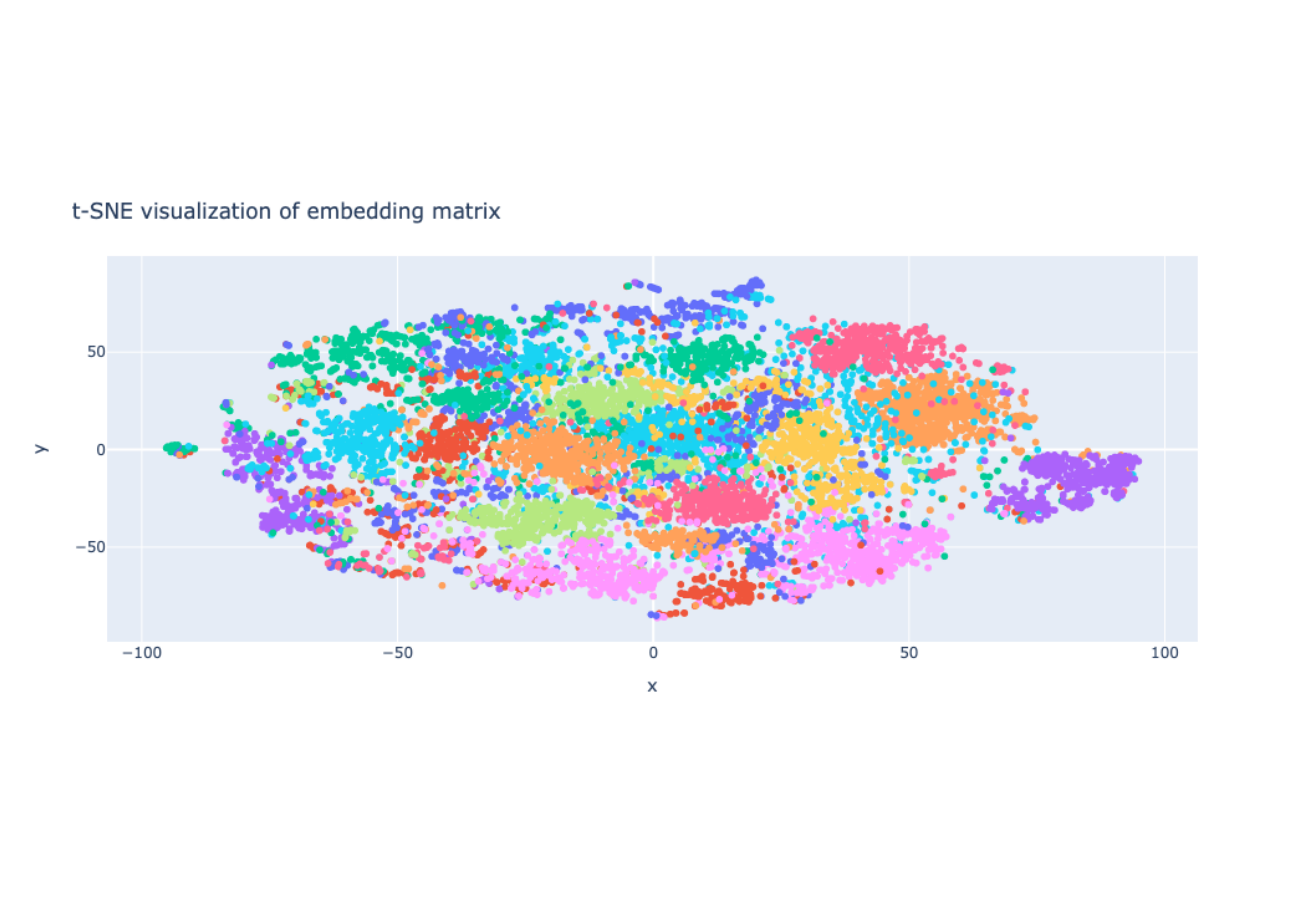}}
  \caption{t-SNE visualization of clustering of chunks in embedding space for Kumar and Clark Clinical Medicine}
  \label{fig:fig3}
\end{figure}

Kumar and Clark had 1508 pages with 13024 text chunks, each having an average of 789 tokens. The maximum affordable token limit was set at 15,000 and 19 representative chunks were selected. The word cloud and the t-SNE visualization are depicted in Figures \ref{fig:fig2} and \ref{fig:fig3}.

\begin{figure}
  \centering
  \fbox{\includegraphics[width=0.5\textwidth]{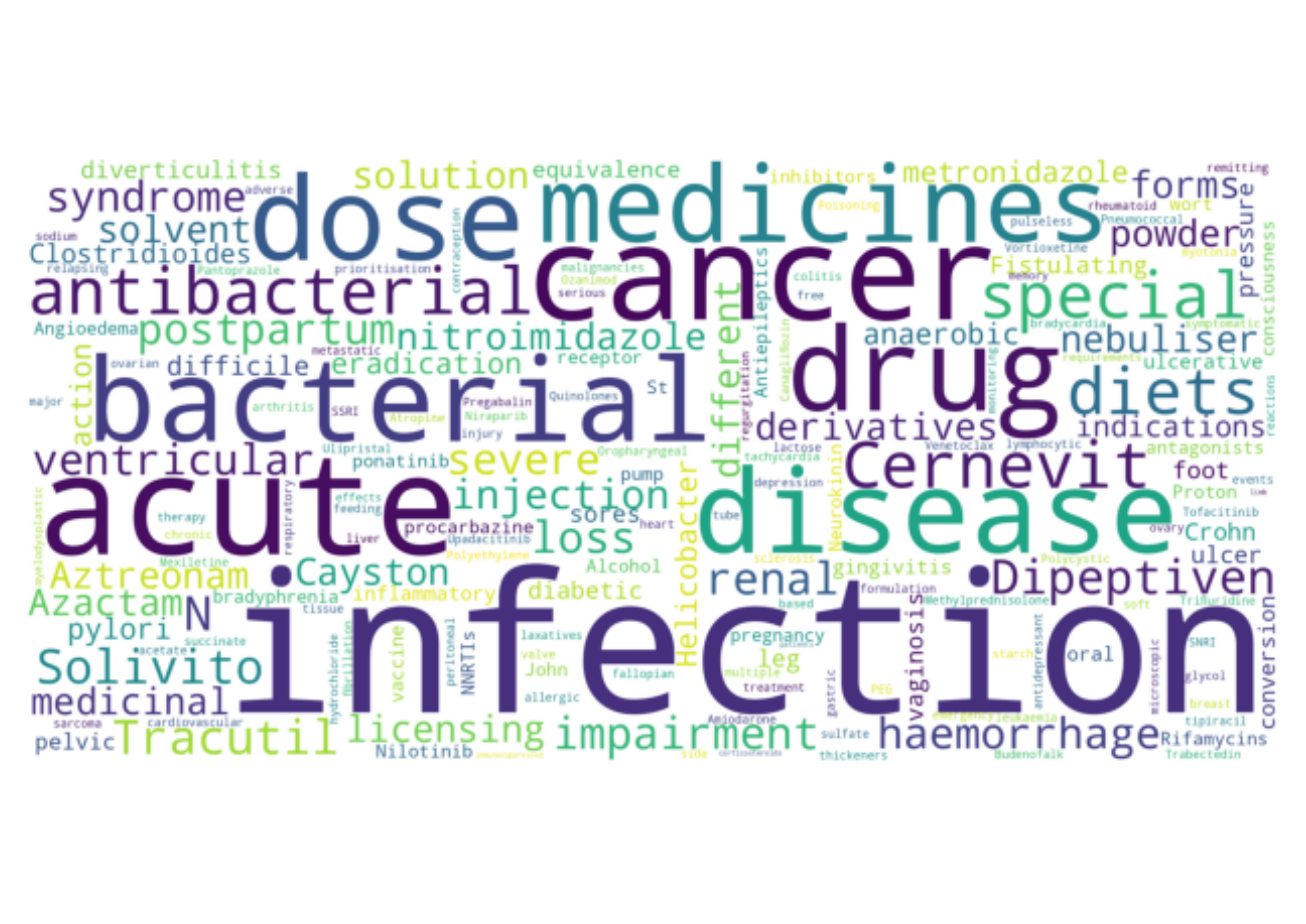}}
  \caption{Word cloud BNF 82}
  \label{fig:fig4}
\end{figure}

\begin{figure}
  \centering
  \fbox{\includegraphics[width=0.5\textwidth]{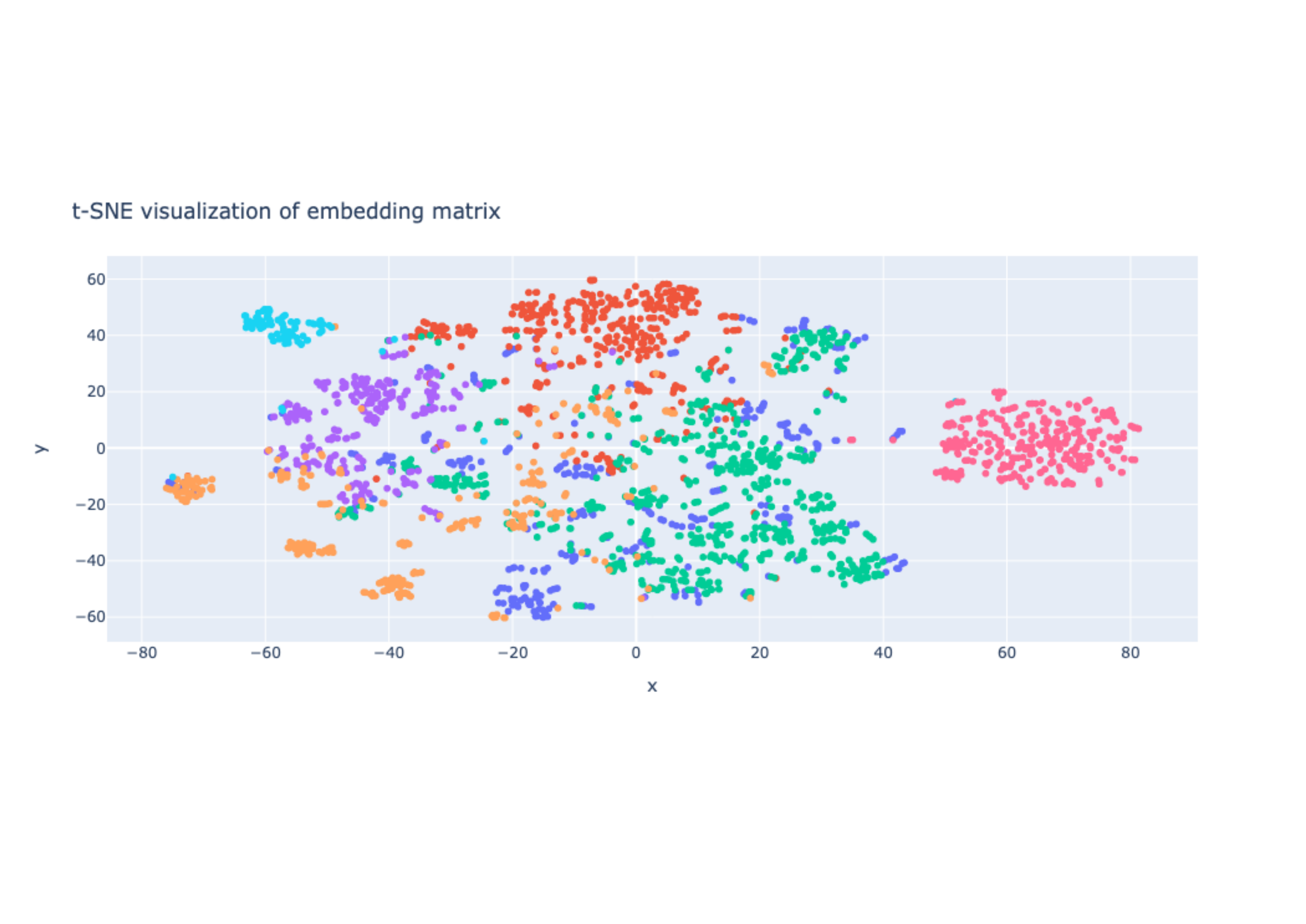}}
  \caption{t-SNE visualization of clustering of chunks in embedding space for BNF}
  \label{fig:fig5}
\end{figure}

BNF had 1805 pages with 7278 text chunks with an average token size of 486. The model chose 10 representative chunks under the constraints of 5000 maximum affordable tokens. The word cloud and t-SNE are depicted in Figures \ref{fig:fig4} and \ref{fig:fig5}.

\section{Implementation}

RVS is implemented in \verb|docGPT|, a document intelligence program written in Python using \verb|langchain| framework and the source is available at \url{https://github.com/ssm123ssm/docGPT-pharm}.

\section{Conclusion}

Clinical medicine is a knowledge-intensive domain. Both clinicians and medical students would benefit from efficient methods for retrieving information quickly from large knowledgebases. We believe the proposed retrieval augmented generation workflow and representative vector summarization for large documents would be of help in this context. Even though the workflow was tested on medical reference books and use cases related to medical education, the concept of RAG and RVS can be adopted by other domains as well.

\bibliographystyle{unsrt}  
\bibliography{references}

\end{document}